\documentclass[runningheads]{llncs}
\usepackage{graphicx}
\usepackage{comment}
\usepackage{amsmath,amssymb} 
\usepackage{color}
\usepackage{lipsum} 
\usepackage{multirow}
\usepackage{array}
\usepackage{hyperref}
\usepackage{footnote}

\DeclareMathOperator*{\argmin}{arg\,min}

\usepackage[width=122mm,left=12mm,paperwidth=146mm,height=193mm,top=12mm,paperheight=217mm]{geometry}

\begin{document}
\pagestyle{headings}
\mainmatter
\def\ECCVSubNumber{1093}  

\title{Semantic Correspondence via 2D-3D-2D Cycle} 


\titlerunning{Semantic Correspondence via 2D-3D-2D Cycle}
\author{Yang You, Chengkun Li, Yujing Lou, Zhoujun Cheng, \\ Lizhuang Ma, Cewu Lu, Weiming Wang\thanks{Weiming Wang is the corresponding author.}}
\authorrunning{Y. You et al.}
%
\institute{Shanghai Jiao Tong University, China\\
\email{\{qq456cvb,sjtulck,louyujing,blankcheng\}@sjtu.edu.cn,\\ \{ma-lz,lucewu,wangweiming\}@sjtu.edu.cn}}
\maketitle

\begin{abstract}
Visual semantic correspondence is an important topic in computer vision and could help machine understand objects in our daily life. However, most previous methods directly train on correspondences in 2D images, which is end-to-end but loses plenty of information in 3D spaces. In this paper, we propose a new method on predicting semantic correspondences by leveraging it to 3D domain and then project corresponding 3D models back to 2D domain, with their semantic labels. Our method leverages the advantages in 3D vision and can explicitly reason about objects self-occlusion and visibility. We show that our method gives comparative and even superior results on standard semantic benchmarks. We also conduct thorough and detailed experiments to analyze our network components. The code and experiments are publicly available at \href{https://github.com/qq456cvb/SemanticTransfer}{https://github.com/qq456cvb/SemanticTransfer}.

\end{abstract}

\section{Introduction}
Semantic correspondence for general objects is an important research area for machine vision. Understanding different objects of the same category is still a challenge topic. There are quite a few methods on solving this problem in 2D image domain. \cite{choy2016universal,han2017scnet,kim2017fcss,horn1981determining,liu2010sift} propose to matching local regions between pairs of images while~\cite{rocco2017convolutional,rocco2018end,seo2018attentive,ham2017proposal} consider it as a global image alignment problem. However, these works all investigate 2D image features and very few works focus on the internal 3D structures of images to be matched. We argue that by explicitly exploiting 3D structures of objects, one can easily infer the self-occlusion and spatial relationships. This idea is explored in some recent works~\cite{kulkarni2019canonical,zhou2016learning}, which come up with a 3D model as an intermediate medium. However, they assume there exists a template model for all images, which does not hold in most cases.

\begin{figure}[ht]
    \centering
    \includegraphics[width=0.8\linewidth]{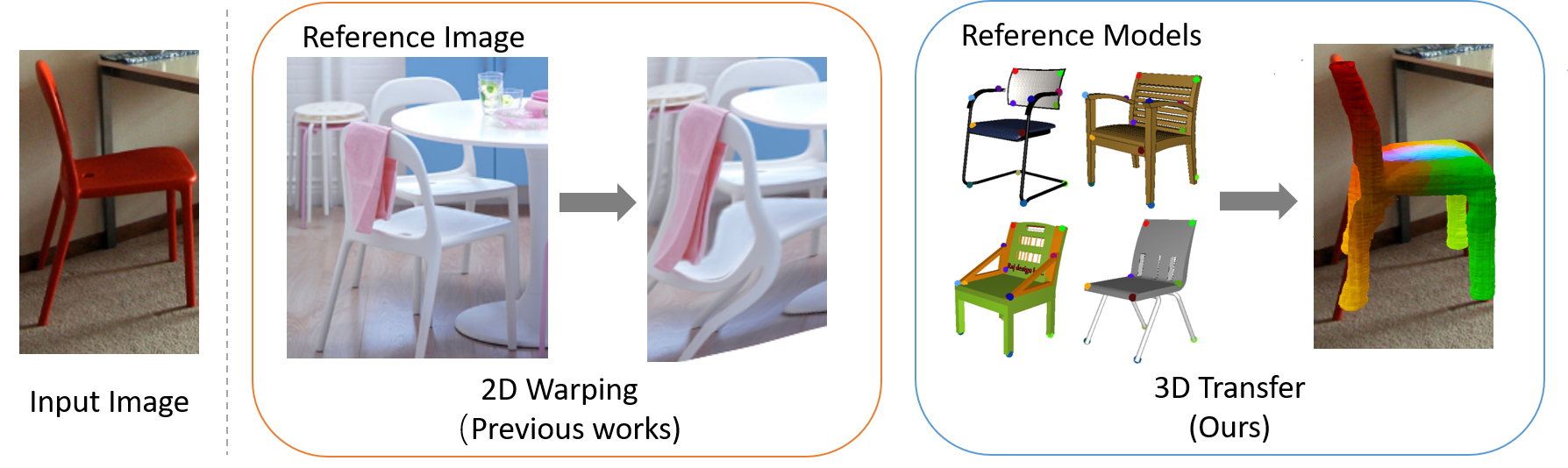}
    \caption{\textbf{Direct 2D-2D correspondence vs. ours 2D-3D-2D pipeline}. Left: direct 2D semantic warping. It is erroneous to directly warp one image to another in 2D domain due to the existence of self-occlusion and viewpoint variations. Right: our method estimates the corresponding 3D model and projects predicted 3D semantic labels back onto 2D images.  }
    \label{fig:intro}
\end{figure}

To solve these problems, we propose a novel semantic transfer method that aims to predict 3D structures from a single RGB image and then project 3D semantic labels back onto 2D image planes. 2D to 3D shape prediction is inspired by Wu et al.~\cite{wu2018learning}. For 3D-2D projection, we estimate viewpoints directly from 2D images and then leverage a 3D semantic prediction model trained on KeypointNet~\cite{you2020keypointnet} to give 3D semantic labels together with its 2D projections. Viewpoints are further fine-tuned with differentiable renderers. The main advantages of this method lies in two aspects. 1) the number of training data required is reduced drastically. Previous 2D image transfer networks require numerous images for a class of object in order to extract robust semantic features. These images view objects of different shapes from different angles. On the contrary, if we could infer 3D structures from 2D images, then all we need is to utilize labels on existing 3D models and then project them onto 2D image planes. One may consider 3D structures inference as a data-heavy task but virtual 2D images can be generated from 3D models on the fly, as done by Wu et al.~\cite{wu2018learning}. 2) Visibility reasoning is made explicit. When we project semantic labels onto 2D image planes, points on the back is naturally culled. On 2D image domains, visibility is implicitly made by 2D CNNs, making it hard to interpret. As shown in Figure~\ref{fig:intro}, for direct 2D-2D methods, the generated 2D warping from source to target does not account for self-occlusion and may be erroneous. However, semantic transfer would be much easier if we first estimate their corresponding 3D models and camera poses.

Although this idea is appealing, there are still a lot of challenges in this 2D-3D-2D cycle. Compared with directly learning correspondence maps from 2D images, our pipeline is more involved. Each stage would incur some error and the final result would get biased through error accumulation. To this end, we propose a camera pose estimation module fine-tuned by differentiable renderer. The final result is competitive and even outperforms state-of-the-art methods on some benchmarks. In addition, we conduct comprehensive and detailed experiments to figure out the effectiveness of each stage in the proposed 2D-3D-2D cycle. We hope this analysis could help future researches to further improve on 2D-3D, 3D-2D or 2D-3D-2D predictions.

Our main contributions are listed below:
\begin{itemize}
    \item We propose a novel 2D-3D-2D pipeline to solve 2D semantic correspondence problem by leveraging it to the 3D domain.
    \item Our proposed method sets a new state of the art on several semantic correspondence benchmarks. 
    \item We conduct detailed and comprehensive experiments to decompose each stage/parts effect on the accuracy of final results.
    \item We will make all our code with detailed experiments publicly available.
\end{itemize}

\section{Related Work}
\subsection{2D Semantic Correspondence}
Image semantic correspondence has a long history which dates back to optical flow~\cite{horn1981determining}, multi-stereo~\cite{okutomi1993multiple}. Recently, some local descriptor based methods like proposal flow~\cite{ham2017proposal} and SIFT flow~\cite{liu2010sift}  are explored to find dense correspondences across different objects. With the advance of deep learning, neural features~\cite{hariharan2015hypercolumns,lin2017feature,kong2016hypernet} are broadly used as they are more robust and generalizable. Methods like A2Net~\cite{seo2018attentive}, NC-Net~\cite{rocco2018neighbourhood} and HPF~\cite{min2019hyperpixel} view semantic correspondence as a matching problem in high-dimensional feature images. In addition, \cite{florence2018dense,schmidt2016self} leverage an unsupervised methods to learn consistent dense embeddings with SLAM across different objects.

\subsection{3D Semantic Correspondence}
\cite{allen2003space,blanz1999morphable} are the pioneers on detecting 3D semantic correspondence between human bodies and faces. Recently, \cite{halimi2018self,roufosse2019unsupervised,groueix20183d} propose unsupervised methods on learning dense correspondences between humans and animals. With the help of recent large scale model dataset such as ShapeNet~\cite{chang2015shapenet} and PartNet~\cite{mo2019partnet}, finding semantic correspondences on general objects become possible. Deep functional dictionaries~\cite{sung2018deep} and SyncSpecCNN~\cite{yi2017syncspeccnn} all learn a set of synchronized base functions in order to obtain dense correspondence from functional maps. In addition to ShapeNet, \cite{pavlakos20176,kim2013learning,you2019fine,you2020keypointnet} provide additional keypoint or correspondence annotations for object semantic understandings. 

Perhaps, CSM~\cite{kulkarni2019canonical} and Zhou et al.~\cite{zhou2016learning} are the closest to this paper. However, they assume that for all images, there is a template 3D model that fits well, making them not directly applicable to categories where the shapes across instances differ significantly in topology or undergo large articulation. Besides, they implicitly infer 3D models by generating a 2D-3D pixel maps while we explicitly predict each image's corresponding 3D shape.

\subsection{Single View Shape Reconstruction} Recently, many works have been introduced on single view shape reconstruction. For supervised methods where a ground-truth model is available, PSGN~\cite{fan2017point} and pseudo-renderer~\cite{lin2018learning} reconstruct point clouds from single-view RGB images. Front2back~\cite{yao2019front2back} predicts per-pixel depth, which is then converted into a point cloud.  \cite{wu2017marrnet,chen2009learning,wu2016learning} predict voxel grids with a relatively small resolution while some others like~\cite{park2019deepsdf,mescheder2019occupancy,liu2019learning} reconstruct implicit surface functions, where resolutions are not limited compared to voxels. In addition, there are also plenty of researches~\cite{gkioxari2019mesh,wen2019pixel2mesh++} focused on triangle mesh reconstruction, which is constrained by mesh topology. Pan et al.~\cite{pan2019deep} tries to modify the mesh topology during reconstruction. What's more, some other directions like reconstructing images as geometric primitive collections~\cite{gao2019sdm,tian2019learning} and complex octree structures~\cite{riegler2017octnet,tatarchenko2017octree} are also explored. 

For unsupervised single view shape prediction, \cite{sitzmann2019deepvoxels,niemeyer2019differentiable,rematas2019neural,eslami2018neural} utilize only 2D image annotations, together with a multi-view consistency prior, to reconstruct the implicit 3D models. Other works like~\cite{wang2020deep,li2019synthesizing} focus on a large collection of images in the wild and reconstruct a model for each distinguished image.

\subsection{Differentiable Renderer}
Differentiable renderer is an emerging topic in recent years, we see that ~\cite{sitzmann2019deepvoxels,niemeyer2019differentiable} all utilize differentiable projections to learn a 3D shape from its 2D image projections. Neural Mesh Renderer~\cite{kato2018neural} first brings this idea to mesh rasterization rendering and Li et al.~\cite{li2018differentiable} comes up with differentiable monte-carlo ray tracing. DiffSDF~\cite{jiang2019sdfdiff} renders implicit surfaces defined by signed distance function in a differentiable way.

\section{Methodology}
\subsection{Overview}
\begin{figure}[ht]
    \centering
    \includegraphics[width=0.8\linewidth]{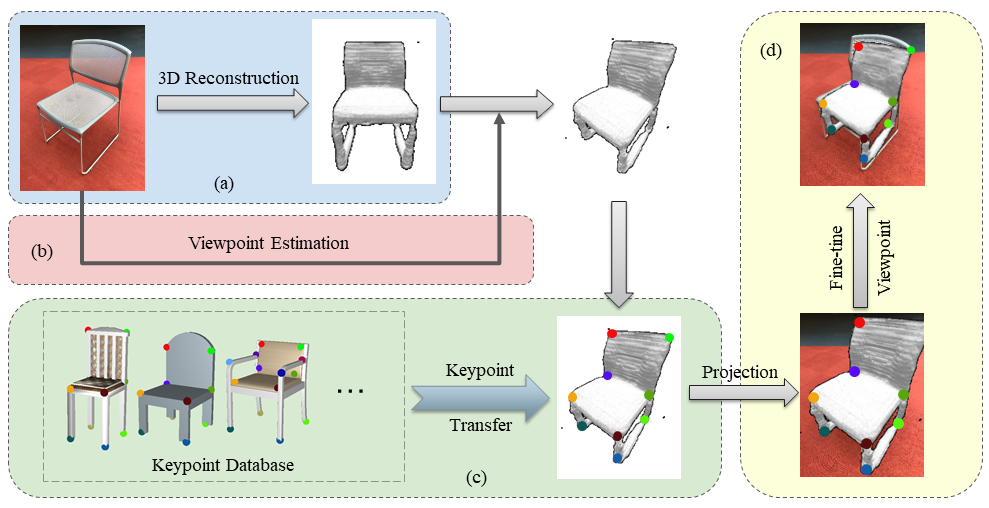}
    \caption{\textbf{Our full pipeline}. (a) 3D models are reconstructed from single RGB images. (b) Viewpoints are also estimated from RGB images. (c) We obtain a keypoint descriptor database by training on existing 3D keypoint datasets, and then transfer these keypoints with nearest neighbor search. (d) Given viewpoints, 3D models and transferred keypoints, we project them onto the original image plane.}
    \label{fig:pipeline}
\end{figure}

Our method includes four essential parts: (a) 2D-3D shape prediction, (b) viewpoint estimation from RGB images, (c) keypoint database with semantic embeddings, (d) 3D-2D projections of semantic points. Full pipeline is illustrated in Figure~\ref{fig:pipeline}.

For 2D-3D shape prediction, we utilize a similar structure with ShapeHD~\cite{wu2018learning}, which first estimate silhouettes, normals and depths from 2D images and then predict 3D shapes using 3D convolutions. This pattern can be summarized as 2D-2.5D-3D and is first proposed by MarrNet\cite{wu2017marrnet}. For viewpoint estimation, azimuth and elevation are classified into discrete bins using ResNet architecture. For 3D keypoint transfer, we utilize KeypointNet~\cite{you2020keypointnet} annotations with nearest neighbor search. For 3D-2D semantic projection, 3D voxel predictions are converted into meshes with marching cube algorithm~\cite{lorensen1987marching}. Then, these meshes can be projected back onto 2D image films provided viewpoint estimations, fine-tuned by differentiable renderer. Note that we assume that objects are centered in the image and do not get occluded or cut by other objects. Clutter occlusions and incompletions are out of the scope of this paper.

\subsection{Single View 3D Reconstruction}
\begin{figure}[ht]
    \centering
    \includegraphics[width=0.8\linewidth]{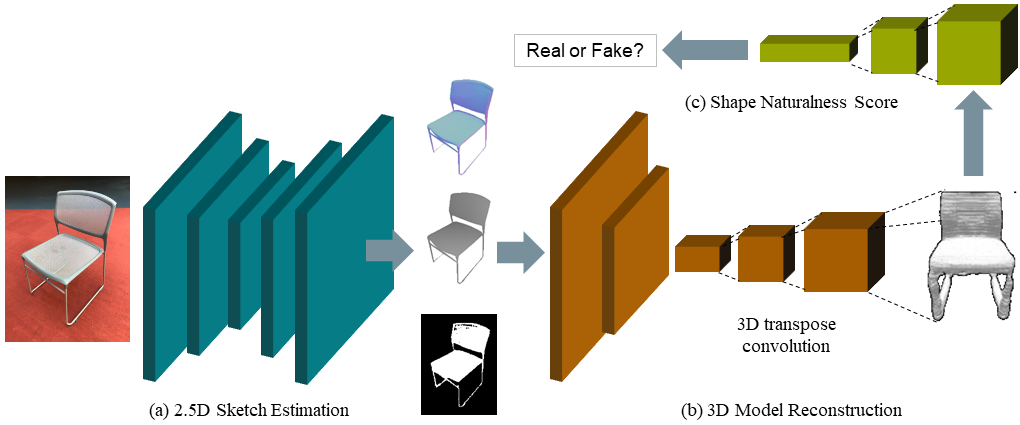}
    \caption{\textbf{Single view 3D model reconstruction}. (a) Firstly, 2.5D sketches including normals, silhouettes and depths are estimated. (b) Then, 3D transpose convolution is used to recover object voxels. (c) In addition, a shape naturalness score is proposed to ensure that generated shapes are not diverged from real shapes.}
    \label{fig:3dreconstruction}
\end{figure}

There are a number of works focusing on single view 3D reconstruction, such as MarrNet~\cite{wu2017marrnet}, ShapeHD~\cite{wu2018learning}, Mesh R-CNN~\cite{gkioxari2019mesh}. We utilize a similar architecture with ShapeHD, considering that it could penalize those unrealistic 3D shapes. To make it complete, here we briefly show the components that are used in ShapeHD. ShapeHD is inspired by MarrNet. It has an 2.5D sketch estimator, which is an encoder-decoder that predicts the object's depth, surface normals and silhouette from an RGB image. Followed is a 3D estimator which also has an encoder-decoder structure. It predicts a 3D shape of the object in the canonical view from 2.5D sketches. In addition, the author introduced a deep naturalness regularizer that penalizes unrealistic shapes prediction. The regularizer is implemented by a 3D generative adversarial network and the discriminator is then used to calculate the naturalness score.
The architecture for this module is shown in Figure~\ref{fig:3dreconstruction}.

\subsection{Viewpoint Estimation from RGB Images}
Given predicted 3D shapes, it is necessary to estimate the viewpoint in order to project it back onto the image plane. To do so, we design a network that predict viewpoints directly from RGB images. Note that this is different from Pix3D~\cite{sun2018pix3d} where viewpoints are estimated from 2.5D sketches. We argue that error would accumulate if the previous 2.5D sketch prediction is inaccurate. Direct estimation reduces the number of passed stage from two to one, which help improve the accuracy. This is also verified in our experiments.

\begin{figure}[ht]
    \centering
    \includegraphics[width=0.8\linewidth]{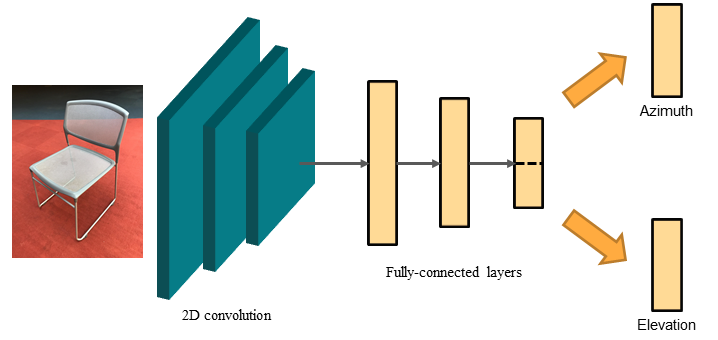}
    \caption{\textbf{Viewpoint Estimation Module}. Viewpoint is estimated from 2D CNN followed by several fully connected layers. Then, KL divergence loss is employed.}
    \label{fig:viewpoint}
\end{figure}

We treat view estimation as a classification problem, where azimuth is divided into 24 bins the elevation is divided into 12 bins. Circularity in azimuth is dealt carefully with an additional circular bin.
The architecture for this module is demonstrated in Figure~\ref{fig:viewpoint}.

\subsection{3D Semantic Prediction}
Predicting semantic labels on 3D models is pretty challenging in this 2D-3D-2D loop. Firstly, the predicted 3D shape from previous stage is not perfect and may be corrupted. Secondly, directly training on 3D models may be prohibitive as current semantic image datasets usually do not come up with the corresponding 3D models. Even for datasets with 3D models like PASCAL-3D~\cite{xiang_wacv14}, since the number of models is relatively small, overfitting is highly suspected. 

\begin{figure}[ht]
    \centering
    \includegraphics[width=0.8\linewidth]{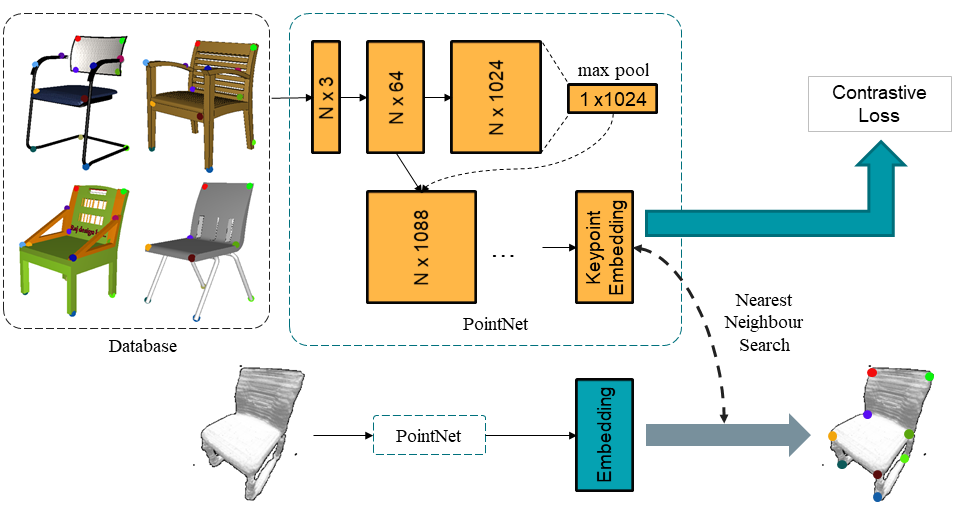}
    \caption{\textbf{Semantic Keypoint Transfer}. The database has a large collection of models (may not necessarily contain the model to be evaluated). We extract their keypoint embeddings using PointNet trained with contrastive loss. For the input model, its dense embeddings are extracted with the same pretrained PointNet. Afterwards, keypoint locations are identified by a nearest neighbor search.}
    \label{fig:semantictransfer}
\end{figure}

Therefore, we resort to existing large-scale 3D keypoint datasets to train a semantic prediction network. KeypointNet\cite{you2020keypointnet} contains millions of keypoint annotations from ShapeNet models. By training on this dataset, one could obtain a semantic embedding for each keypoint in the dataset and then transfer it to the predicted 3D models with a nearest neighbor search. In other words, we train a semantic prediction network on a 3D object database and then generalize it to our predicted 3D shapes. To account for corruption, we augment our dataset with random Gaussian noises near the object surface.
This idea is illustrated in Figure~\ref{fig:semantictransfer}.

\subsection{Differentiable Rendering in 2D Projection}
As a final step, we are now ready to project our predicted 3D shapes together with inferred semantic points back onto 2D image planes. This step, although the last but not the least, is important as errors are accumulated all the way through previous stages. To have a chance correcting previous predictions, we threshold the voxels with marching cube algorithm, fine-tuning the viewpoint with the help of Neural Mesh Renderer~\cite{kato2018neural}. Specifically, denote the ground-truth silhouette image as $S$, predicted 3D model as $M$, our fine-tuned viewpoint is $V$:

\begin{align}
    V^* = \argmin_V\sum_{p\in \Omega}(S[p] - \mathrm{Proj}(M, V)[p]),
\end{align}

where $\Omega$ is the set of all 2D image coordinates, $\mathrm{Proj}(M,V)$ is the projected image given 3D model $M$ under viewpoint $V$. We run several gradient descent steps in order to find the best viewpoint.

To summarize, we first predict 3D shapes and viewpoints from single view RGB images; then 3D semantic keypoints or any other semantic information is transferred from an existing 3D model database to the predicted 3D shapes; finally, the predicted 3D shapes together with their semantics are projected onto 2D image planes, with viewpoints fine-tuned.

\begin{figure}[ht]
    \centering
    \includegraphics[width=0.8\textwidth]{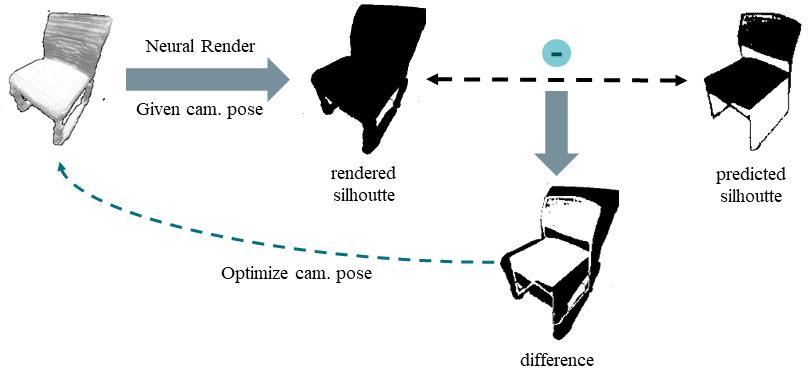}
    \caption{\textbf{Differentiable Rendering}. Given current camera pose/viewpoints, we back-propagate through neural mesh renderer to optimize its pose by comparing rendered silhouette and predicted silhoutte.}
    \label{fig:neuralrender}
\end{figure}





\section{Experiments}
Our experiments are divided into three parts. The first part is the comparison of our proposed method with current state-of-the-art methods. The second part is some ablation studies on our proposed viewpoint estimation and fine-tuning modules.   The third part is a detailed and thorough investigation of each component/stage's influence on final results in our full pipeline. We hope this kind of detailed analysis could help following researchers to have a better understanding on each individual components in the 2D-3D-2D loop. Note this analysis also covers previous 2D-2.5D-3D reconstruction pipeline and can be used to further improve single-view 3D reconstruction.

\paragraph{Datasets}
\begin{figure}
    \centering
    \includegraphics[width=0.8\linewidth]{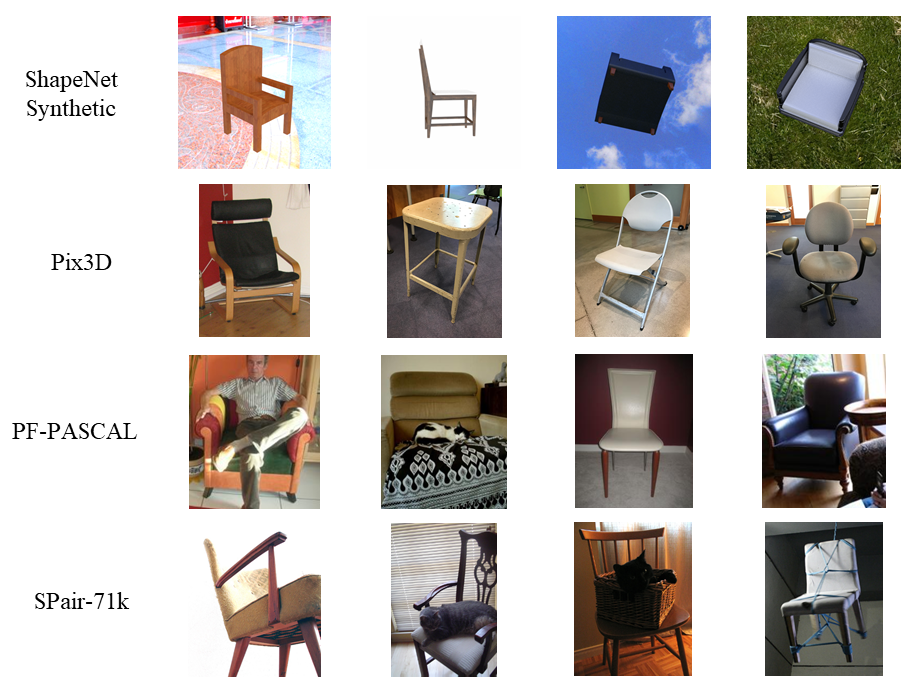}
    \caption{\textbf{Dataset Visualization (chair) on ShapeNet, Pix3D, PF-PASCAL and SPair-71k}. From up to bottom: difficulties from easy to hard. }
    \label{fig:discrep}
\end{figure}

We use ShapeNet Synthetic rendered images~\cite{wu2018learning} for training and Pix3D~\cite{sun2018pix3d}, ~PF-PASCAL\cite{ham2016proposal}, SPair-71k~\cite{min2019spair} images for evaluation (except for Section~\ref{sec:domaingap}). Here, chairs are chosen for better illustrations. More results on other classes can be found in the supplementary material. Differences among these four datasets are shown in Figure~\ref{fig:discrep}. ShapeNet Synthetic could provide tons of training data though it is fake and synthetically rendered. Pix3D is a real dataset without much clutter occlusions and objects are well centered in images, which is relatively clean. PF-PASCAL and SPair-71k provide more extreme occlusions/cutoff/scale variations. Though clutter occlusions and incompletions are not the focus of this paper, our method still gives a competitive score on these datasets compared with state-of-the-art.

\paragraph{Metric}
We use a common evaluation metric of percentage of correct keypoints (PCK), which counts the average number of correctly predicted keypoints given a tolerance threshold. Given predicted keypoint $\mathbf{k}_{pr}$ and groundtruth keypoint $\mathbf{k}_{gt}$, the prediction is considered correct if
Euclidean distance between them is smaller than a given
threshold. The correctness $c$ of each keypoint can be expressed as

\begin{align}
    c = \begin{cases}
1 &\text{if\quad$d(\mathbf{k}_{pr},\mathbf{k}_{gt})\le \alpha_\tau\cdot\mathrm{max}(w_\tau,h_\tau)$}\\
0 &\text{otherwise,}
\end{cases}
\end{align}
where $w_\tau$ and $h_\tau$ are the width and height of either an entire image or object bounding box, $\tau\in$\{img, bbox\}, and $\alpha_\tau$ is a tolerance factor.

We select out those keypoints that are in the intersection of both KPNet and Pix3D/PF-PASCAL/SPair-71k for evaluation. For our method, we directly estimate keypoints for each single input image and calculate PCK for each keypoint; while for other baseline methods like Hyperpixel Flow, PCK is calculated on an image pair where  keypoints in one image are reckoned as ground-truth and keypoints in the other image are predicted by a semantic warp or transfer. Our method can be also considered as transferring keypoints from an implicit ground-truth keypoint template. PCK results are averaged over all keypoints. 

All our networks are written in Pytorch. Each stage in Figure~\ref{fig:pipeline} is trained independently. All input images are cropped and resized to $480\times 480$ so that the object is centered in the image.


\subsection{Comparison with State-of-the-Arts}
\label{sec:soa}
In this section, we compare our methods with several state-of-the-arts that either directly do a 2D image semantic transfer~\cite{min2019hyperpixel,seo2018attentive} or utilize 3D templates~\cite{kulkarni2019canonical}. 

\begin{figure}[ht]
    \centering
    \includegraphics[width=0.8\linewidth]{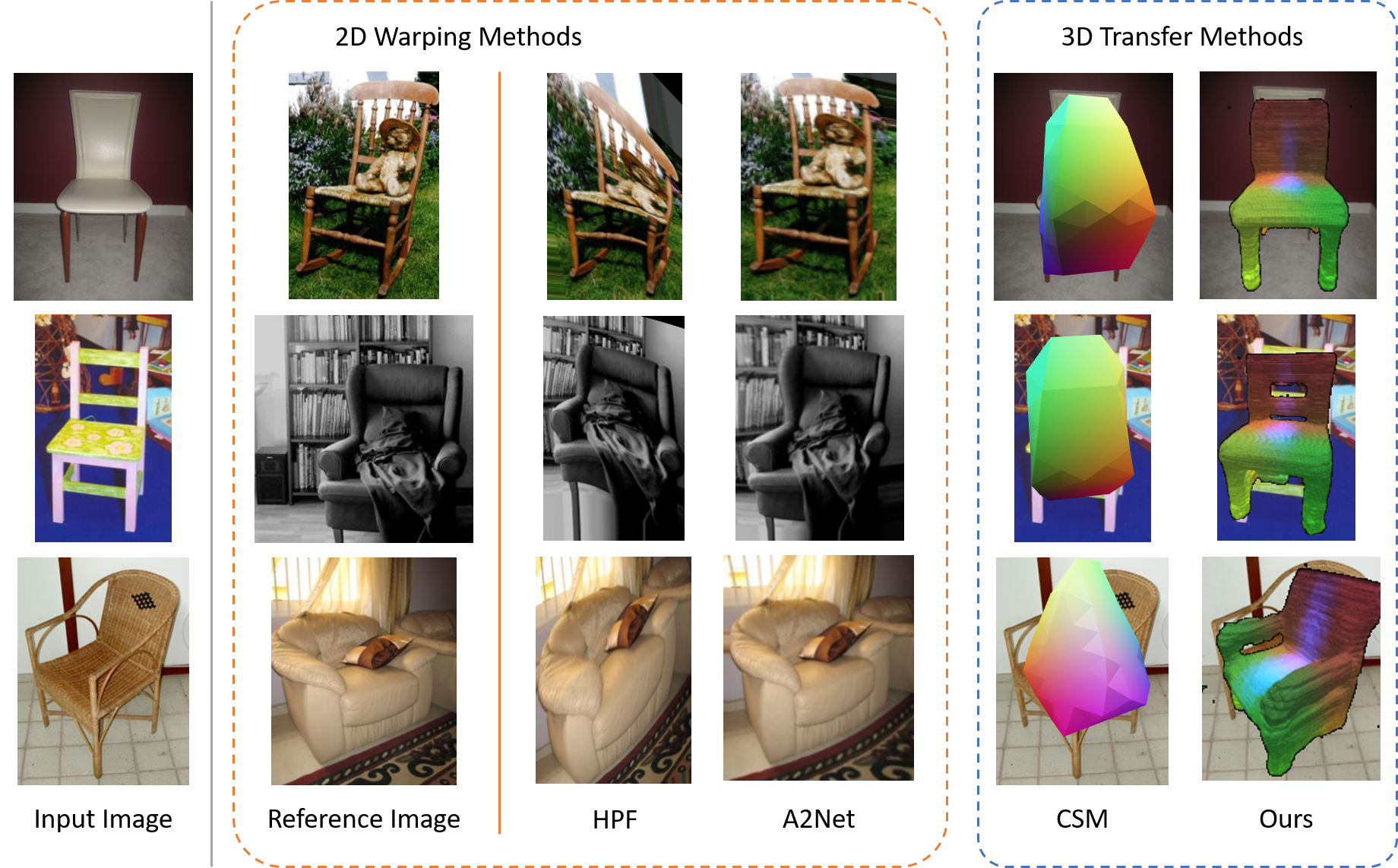}
    \caption{\textbf{Qualitative results on PF-PASCAL}.}
    \label{fig:pfpascal}
\end{figure}

\begin{figure}[ht]
    \centering
    \includegraphics[width=0.8\linewidth]{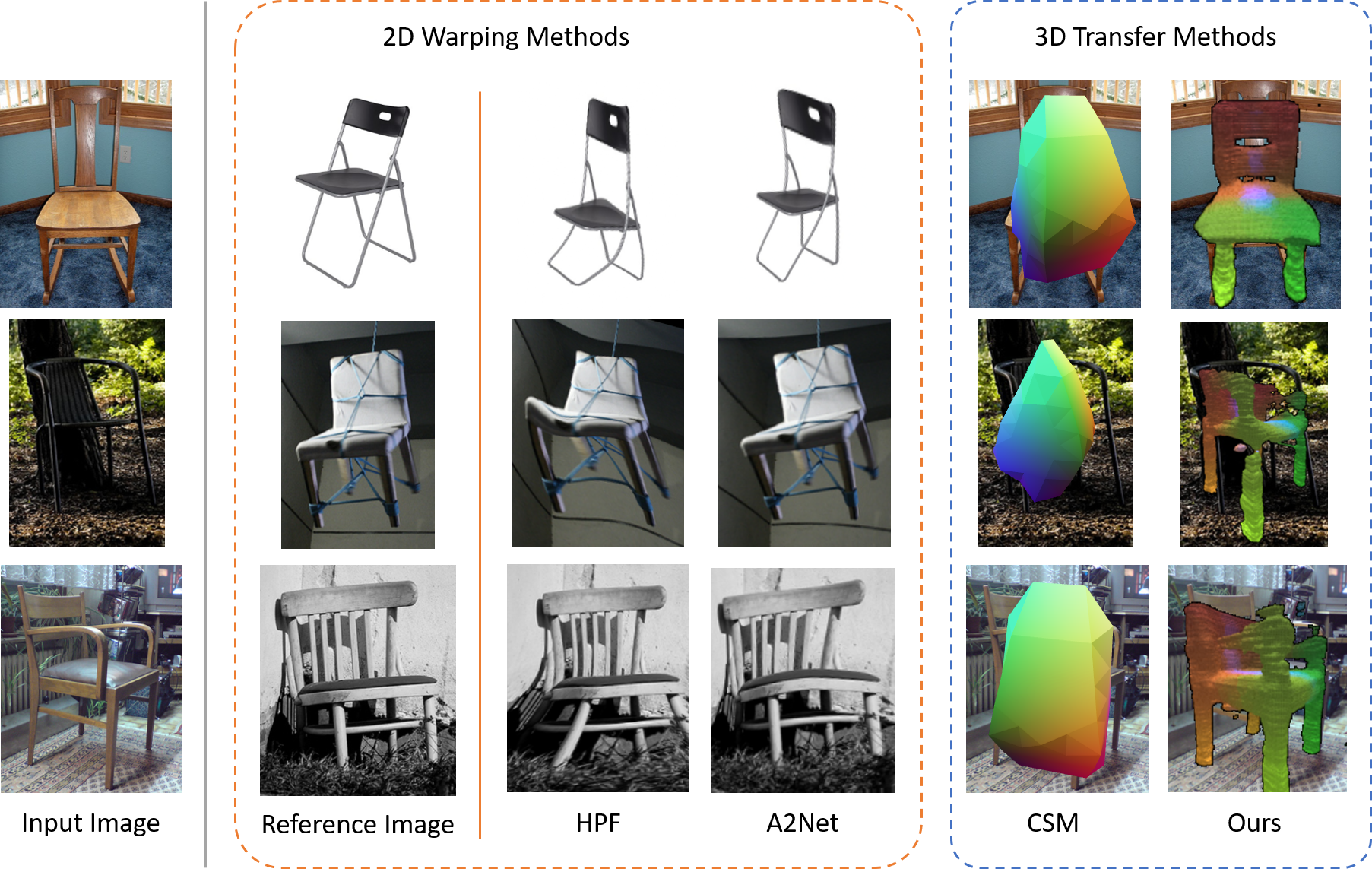}
    \caption{\textbf{Qualitative results on SPair-71k}.}
    \label{fig:spair}
\end{figure}

Our method is trained with ShapeNet Synthetic renderings while state-of-the-art methods are trained directly on real-world images. Interestingly, though our method has a domain gap when applied to real-world images, we still outperform state-of-the-art methods on Spair-71k and Pix3D. 
Quantitative results are shown in Table~\ref{tab:compsoa}. On PF-PASCAL, our method is inferior due to the difficulty in handling severely occluded and incomplete objects. Qualitative results on SPair-71k/PF-PASCAL are shown in Figure~\ref{fig:pfpascal} and \ref{fig:spair}.

\begin{savenotes}
\begin{table}[ht]
\centering
\resizebox{0.85\textwidth}{!}{
\begin{tabular}{l|c|c|c|c|c|c}
    \hline
    \multirow{2}{*}{Models} & \multicolumn{3}{c|}{PCK-chair ($\alpha_{img}=0.1$)} & \multicolumn{3}{c}{PCK-chair ($\alpha_{bbox}=0.1$)} \\
        \cline{2-7}
        & PF-PASCAL & SPair71k & Pix3D & PF-PASCAL & SPair71k & Pix3D\\
    \hline
    ours & 0.444 & \textbf{0.565} & \textbf{0.560} & 0.342 & \textbf{0.346} & 0.323\\
    HPF\textsubscript{res101}\cite{min2019hyperpixel} &  \textbf{0.602}  & 0.419 & 0.534 & \textbf{0.435} & 0.324 & \textbf{0.434}\\
    A2Net\textsubscript{res101}\cite{seo2018attentive} & 0.516 & 0.358 & 0.506 & 0.302 & 0.200 & 0.356\\
    CSM\textsubscript{unet}\cite{kulkarni2019canonical} & 0.164 & 0.115 &0.152 & 0.164\footnotemark[1] & 0.115\footnotemark[1] & 0.152\footnotemark[1]\\
    \hline
\end{tabular}}
\bigbreak
\caption{\textbf{Comparison of our method with state-of-the-arts}. \textsuperscript{1}CSM crops input images with object bounding boxes, so the results for $\alpha_{img}=0.1$ and $\alpha_{bbox}=0.1$ are the same.}
\label{tab:compsoa}
\end{table}
\end{savenotes}


\subsection{Ablation Study on Viewpoint Estimation}
In this section, we explore the effect of proposed view estimation module and viewpoint fine-tuning module. 

\begin{figure}[ht]
    \centering
    \includegraphics[width=0.8\linewidth]{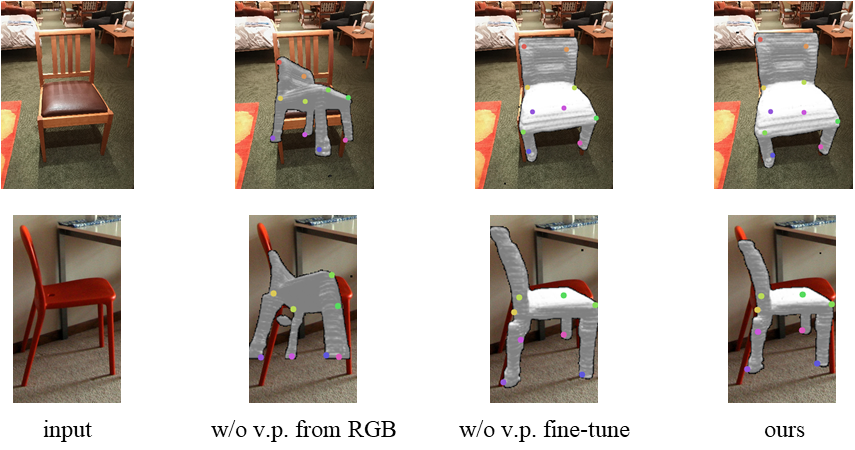}
    \caption{\textbf{Visualization on viewpoint estimation modules}. From left to right: input image; viewpoint estimated from predicted 2.5D sketches instead of RGB images; viewpoint not fine-tuned; models with all modules.}
    \label{fig:ablation}
\end{figure}

For the view estimation module, we compare with viewpoints that are predicted from estimated 2.5D sketch (w/o v.p. from RGB). Quantitative results are shown in Table~\ref{tab:ablation}, we see that our proposed method greatly improve the accuracy of view estimation by not accumulating the error in the 2D-2.5D sketch prediction. From Figure~\ref{fig:ablation}, it can be concluded that viewpoints estimated from 2.5D sketch are much more biased and reduce the quality of final transferred keypoints.

For the viewpoint fine-tuning module, fine-tuning viewpoints improves the overall accuracy on SPair-71k and Pix3D while downgrades on PF-PASCAL when $\alpha_{img}=0.1$, as shown in Table~\ref{tab:ablation}. This is due to the fact that PF-PASCAL includes more occluded chairs than SPair-71k thus is harder than the latter one. This breaks the prior of objects centered on the image thus making viewpoint fine-tuning vulnerable. Qualitative results are given in Figure~\ref{fig:ablation}.

\begin{table}[ht]
\centering
\resizebox{0.9\textwidth}{!}{
\begin{tabular}{l|c|c|c|c|c|c}
    \hline
    \multirow{2}{*}{Models} & \multicolumn{3}{c|}{PCK-chair ($\alpha_{img}=0.1$)} & \multicolumn{3}{c}{PCK-chair ($\alpha_{bbox}=0.1$)} \\
        \cline{2-7}
        & PF-PASCAL & SPair71k & Pix3D & PF-PASCAL & SPair71k & Pix3D\\
    \hline
    ours & 0.444 & \textbf{0.565} & \textbf{0.560} & \textbf{0.342} & \textbf{0.346} & \textbf{0.323}\\
    ours w/o v.p. from RGB  & 0.171 & 0.185 & 0.238 & 0.075 & 0.098 & 0.117\\
    ours w/o v.p. fine-tune & \textbf{0.497} & 0.538 & 0.560 & 0.316 & 0.332 & 0.322\\
    \hline
\end{tabular}}
\bigbreak
\caption{\textbf{Ablation study on viewpoint estimation modules}.}
\label{tab:ablation}
\end{table}

\begin{figure}[h!]
    \centering
    \includegraphics[width=0.8\linewidth]{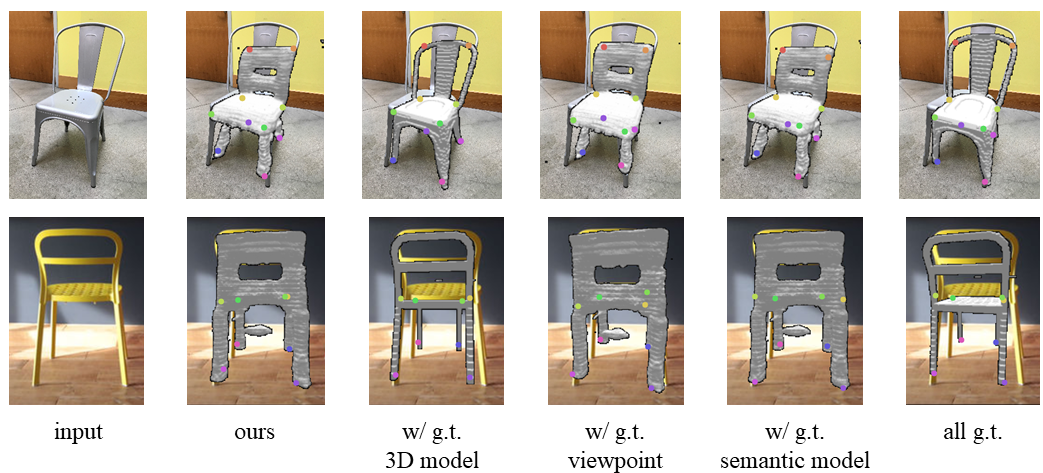}
    \caption{\textbf{Visualization of each stage's effect on Pix3D}. From left to right: input image; replaced with ground-truth 3D model; replaced with ground-truth viewpoint; replaced with ground-truth 3D model in semantic transfer; all ground-truths.}
    \label{fig:gtpix3d}
\end{figure}


\subsection{Detailed Analysis of Each Stage}
In this section, we investigate how each stage influences the final result, by replacing the following three components with their ground-truths: (a) 2D to 3D shape reconstruction, (b) viewpoint estimation and (c) semantic 3D model. Here, semantic 3D model means whether to use ground-truth 3D model of the input image when doing keypoint transfer (the database input to PointNet in Figure~\ref{fig:semantictransfer}).

We start with our full pipeline and then replace each component with its ground-truth counterpart to see the accuracy improvement, respectively. We also evaluate our method with all components' corresponding ground-truths. Notice that although ground-truth viewpoints with azimuth and elevation are given, they are not the ground-truth 6D camera poses. Therefore, all components with ground-truth fails to achieve 100\% accuracy.

Results are given in Table~\ref{tab:gtpix3d}. We see that the PCK contribution of ground-truth semantic 3D model is the largest, which means that if we have the ground-truth model for computing keypoint embeddings instead of the ones in our keypoint database, we would gain about 15.5\% relative improvement. The contribution of ground-truth 3D reconstruction is small, which suggests that the 2D-2.5D-3D single view reconstruction pipeline meets few difficulties when applied to real datasets. Replacing predicted viewpoints with ground-truths also gives 8.2\% PCK improvement, which means that there is still some future work to do in single view camera pose estimation. More visualization results are demonstrated in Figure~\ref{fig:gtpix3d}.

\begin{table}[ht]
\centering
\resizebox{0.9\textwidth}{!}{
\begin{tabular}{l|c|c|c|c}
    \hline
    \multirow{2}{*}{Models} & \multicolumn{2}{c|}{PCK-chair ($\alpha_{img}=0.1$)} & \multicolumn{2}{c}{PCK-chair ($\alpha_{bbox}=0.1$)} \\
        \cline{2-5}
         & Pix3D & ShapeNet & Pix3D & ShapeNet\\
    \hline
    ours &  0.560 & 0.323 & 0.513 & 0.243\\
    ours w/ 2.5D model GT & - & - & 0.518 & 0.243\\
    ours w/ 3D model GT  & 0.571 & 0.363 & 0.521 & 0.263\\
    ours w/ viewpoint GT  & 0.606 & 0.351 & 0.545 & 0.262 \\
    ours w/ semantic model GT & 0.647 & 0.434 & 0.631 & 0.331 \\
    \hline
    all GT & \textbf{0.741} & \textbf{0.560} & \textbf{0.723} & \textbf{0.419}  \\
    \hline
\end{tabular}}
\bigbreak
\caption{\textbf{Detailed analysis of each stage on Pix3D and ShapeNet}.}
\label{tab:gtpix3d}
\end{table}

\subsection{Domain Gap Exploration}
\label{sec:domaingap}
Plenty of single view 3D reconstructions are done on virtually rendered datasets and validated on real-world data. This greatly reduces the need for human annotated 2D keypoint and 3D model label pairs. However, this also introduces an unavoidable gap since due to the rendering error.

In this section, we evaluate our method on virtually rendered ShapeNet, which is from the same domain of training datasets while all the evaluated images are not seen during training. This is in comparison with Section~\ref{sec:soa} whose evaluation is on real datasets. Note, since we are evaluating on rendered datasets, we have 2.5D ground-truth (silhouette, normal and depth), so that we add an extra experiment by replacing 2.5D predictions with ground-truth data.

\begin{figure}[h]
    \centering
    \includegraphics[width=0.8\linewidth]{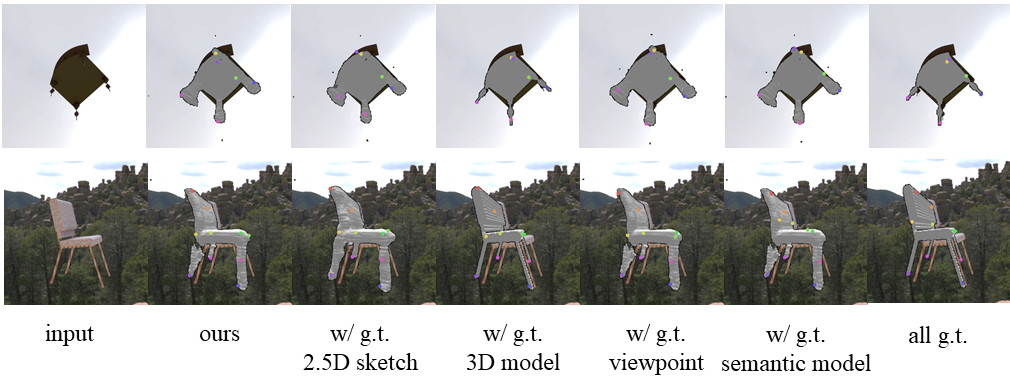}
    \caption{\textbf{Detailed analysis of each stage on ShapeNet Synthetic}. From left to right: input image; replaced with ground-truth 2.5D sketch; replaced with ground-truth 3D model; replaced with ground-truth viewpoint; replaced with ground-truth 3D model in semantic transfer; all ground-truths.}
    \label{fig:gtshapenet}
\end{figure}

Quite interestingly, results on Pix3D are better than on ShapeNet Synthetic. One reason is that in ShapeNet Synthetic dataset, camera positions are sampled uniformly from the entire unit sphere while real datasets seldom have large elevations, making generalization from virtual datasets easier. Quantitative results are given in Table~\ref{tab:gtpix3d} and visualization is shown in Figure~\ref{fig:gtshapenet}.

\begin{figure}[h!]
    \centering
    \includegraphics[width=0.8\linewidth]{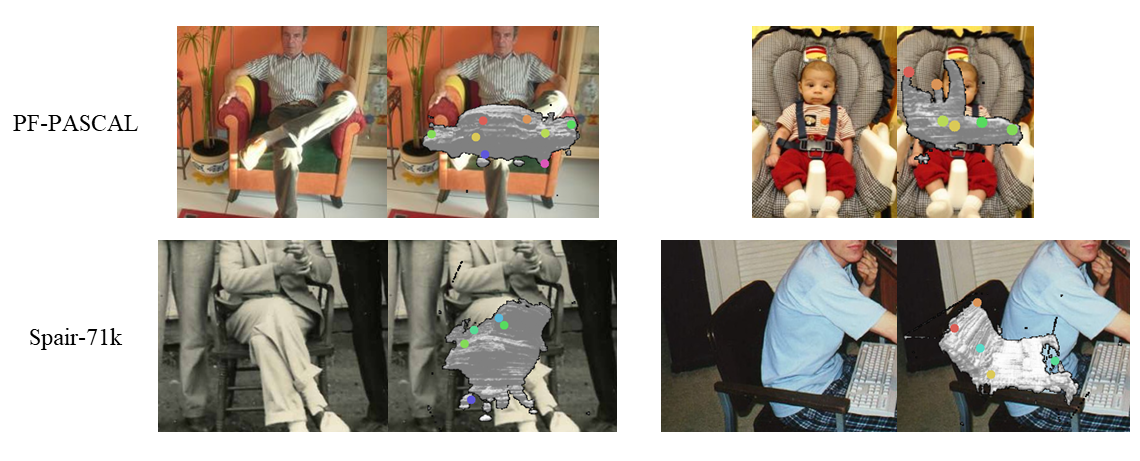}
    \caption{Failure cases of our results on SPair-71k/PF-PASCAL when there exists severe clutter occlusions.}
    \label{fig:failure}
\end{figure}

\section{Future Work}
Here we show some failure cases of our method in Figure~\ref{fig:failure}, where severe clutter occlusions are introduced. It would be interesting to extend our method to explicit reason about clutter occlusions and incompletions. Besides, as our current viewpoint estimation only includes two degrees of freedom, a full 6D pose estimation is possible and we leave it as a future work.

\section{Conclusions}

In this paper, we propose a new pipeline on predicting semantic correspondences by leveraging it to 3D domain and then project corresponding 3D models back to 2D domain, with their semantic labels. This method explicitly reasons about objects self-occlusion and visibility. We show that our method gives comparative and even superior results on standard semantic benchmarks. We also conduct thorough and detailed experiments to analyze our network components.

\clearpage
\section*{Supplementary}

\section*{More Dense Embedding Prediction}
\begin{figure}
    \centering
    \includegraphics[width=\linewidth]{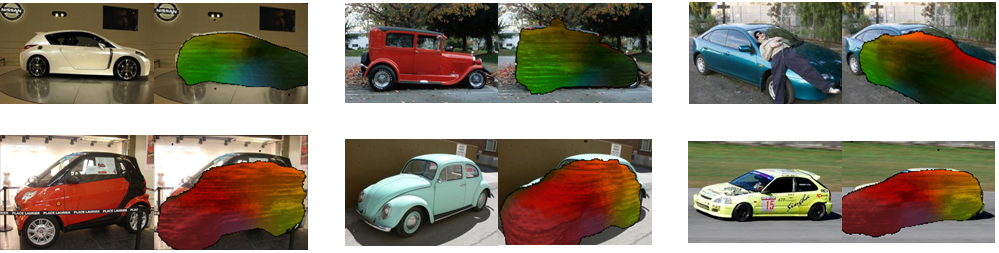}
    \caption{Predicted dense embeddings on PF-PASCAL cars. Notice how the generated embeddings are consistent across different models, despite of viewpoint variations. Similar colors indicate similar embeddings.}
    \label{fig:densecar}
\end{figure}

\begin{figure}
    \centering
    \includegraphics[width=\linewidth]{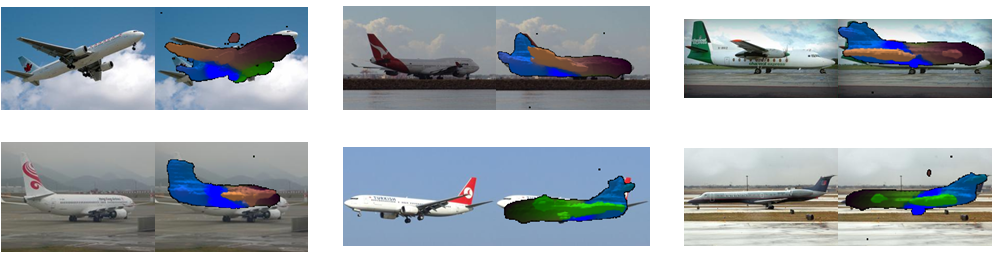}
    \caption{Predicted dense embeddings on PF-PASCAL aeroplanes. Notice how the generated embeddings are consistent across different models, despite of viewpoint variations. Similar color indicate similar embeddings.}
    \label{fig:denseplane}
\end{figure}

In this section, more categories of dense embeddings are visualized in \ref{fig:densecar} and \ref{fig:denseplane}.

\section*{Keypoint Transfer Results on Other Classes}
In this section, we provide additional PCK results on both cars and aeroplanes, evaluated on PF-PASCAL and SPair-71k, in Table~\ref{tab:carsupp} and \ref{tab:planesupp}.
\begin{table}[ht]
    
\centering
\begin{tabular}{l|c|c||c|c}
    \hline
    \multirow{2}{*}{Models} & \multicolumn{2}{c||}{PCK-car ($\alpha_{img}=0.1$)} & \multicolumn{2}{c}{PCK-car ($\alpha_{bbox}=0.1$)} \\
        \cline{2-5}
        & PF-PASCAL & SPair71k & PF-PASCAL & SPair71k \\
    \hline
    ours & \textbf{0.574} & \textbf{0.500} & 0.349 & \textbf{0.328}\\
    HPF\textsubscript{res101} &  0.533  & 0.364  &   \textbf{0.437} & 0.276 \\
    A2Net\textsubscript{res101} &  0.478  &  0.332 &  0.326  & 0.201 \\
    CSM\textsubscript{unet} &  0.339  &  0.234  &  0.339 &   0.234 \\
    \hline
\end{tabular}
\bigbreak
\caption{Comparison of our method with state-of-the-arts on cars.}
\label{tab:carsupp}
\end{table}


\begin{table}[ht]
    
\centering
\begin{tabular}{l|c|c||c|c}
    \hline
    \multirow{2}{*}{Models} & \multicolumn{2}{c||}{PCK-aeroplane ($\alpha_{img}=0.1$)} & \multicolumn{2}{c}{PCK-aeroplane ($\alpha_{bbox}=0.1$)} \\
        \cline{2-5}
        & PF-PASCAL & SPair71k & PF-PASCAL & SPair71k \\
    \hline
    ours & \textbf{0.440} & \textbf{0.390} & 0.194 & 0.182\\
    HPF\textsubscript{res101} &  0.401  & 0.280  &   \textbf{0.294} & \textbf{0.212}\\
    A2Net\textsubscript{res101} &  0.388  & 0.234  &  0.256  & 0.155 \\
    CSM\textsubscript{unet} &  0.220  &   0.139 &  0.220  &   0.139 \\
    \hline
\end{tabular}
\bigbreak
\caption{Comparison of our method with state-of-the-arts on aeroplanes.}
\label{tab:planesupp}
\end{table}




%
%
\bibliographystyle{splncs04}
\bibliography{egbib}
\end{document}